\documentclass[10pt,twocolumn,letterpaper]{article}

\usepackage{graphicx}  
\usepackage[pagenumbers]{cvpr} 
\definecolor{cvprblue}{rgb}{0.21,0.49,0.74}
\usepackage[pagebackref,breaklinks,colorlinks,allcolors=cvprblue]{hyperref}

\def\paperID{5321} 
\def\confName{CVPR}
\def\confYear{2025}
\newsavebox\CBox
\def\textBF#1{\sbox\CBox{#1}\resizebox{\wd\CBox}{\ht\CBox}{\textbf{#1}}\kern-0.05em}
\newcommand{\BEST}[1]{\textbf{\textcolor[rgb]{1.00,0.00,0.00}{#1}}}
\newcommand{\SBEST}[1]{\textbf{\textcolor[rgb]{0.00,0.00,1.00}{#1}}}
\newcommand{\TBEST}[1]{\textbf{\textcolor[rgb]{0.00,1.00,0.00}{#1}}}
\usepackage{color}
\usepackage{amsmath}
\usepackage{graphicx}
\usepackage{subcaption}
\usepackage{multirow}
\usepackage{booktabs} 
\usepackage{tabularx} 
\usepackage{multirow}
\usepackage{arydshln}
\usepackage{algorithm}
\usepackage{diagbox}
\usepackage{algpseudocode}  
\usepackage{amsmath}
\usepackage{flushend}
\usepackage{balance}

\usepackage{stfloats}
\title{Cross-Dataset Gaze Estimation by Evidential Inter-intra Fusion}

\def\spaces{~~~~~~}
\author{Shijing Wang\textsuperscript{1}\spaces{}Yaping Huang\textsuperscript{1}\thanks{Corresponding authors.}\spaces{}Jun Xie$^{2}$\spaces{}Yi Tian\textsuperscript{1}\spaces{}Feng Chen\textsuperscript{2}\spaces{}Zhepeng Wang\textsuperscript{2}\\\\
\textsuperscript{1}Beijing Key Laboratory of Traffc Data Analysis and Mining, Beijing Jiaotong University, China \\
\textsuperscript{2}Lenovo Research
}

\begin{document}
\maketitle
\begin{abstract}
Achieving accurate and reliable gaze predictions in complex and diverse environments remains challenging. 
Fortunately, it is straightforward to access diverse gaze datasets in real-world applications. We discover that training these datasets jointly can significantly improve the generalization of gaze estimation, which is overlooked in previous works.
However, due to the inherent distribution shift across different datasets, simply mixing multiple dataset decreases the performance in the original domain despite gaining better generalization abilities. To address the problem of ``cross-dataset gaze estimation'', we propose a novel \textbf{E}vidential \textbf{I}nter-intra \textbf{F}usion (\textbf{EIF}) framework, for training a cross-dataset model that performs well across all source and target domains. 
Specifically, we build independent single-dataset branches for various datasets where the data space is partitioned into overlapping subspaces within each dataset for local regression, and further create a cross-dataset branch to integrate the generalizable features from single-dataset branches. Furthermore, evidential regressors based on the Normal and Inverse-Gamma (NIG) distribution are designed to additionally provide uncertainty estimation apart from predicting gaze. 
Building upon this foundation, our proposed framework achieves both intra-evidential fusion among multiple local regressors within each dataset and inter-evidential fusion among multiple branches by 
\textbf{M}ixture \textbf{o}f \textbf{N}ormal \textbf{I}nverse-\textbf{G}amma (\textbf{MoNIG}) distribution.
Experiments demonstrate that our method consistently achieves notable improvements in both source domains and target domains. 
\end{abstract}    
\section{Introduction}
Gaze estimation is an important task in computer vision used to determine where a person is looking based on visual cues. It has gained attention for its wide range of applications in human-computer interaction~\cite{majaranta2014eye, rahal2019understanding}, virtual reality~\cite{patney2016perceptually, kim2019nvgaze}, and assistive technology~\cite{jiang2017learning, liu2016identifying, dias2020gaze}. 

\begin{figure}
 \centering
 \includegraphics[width=7cm]{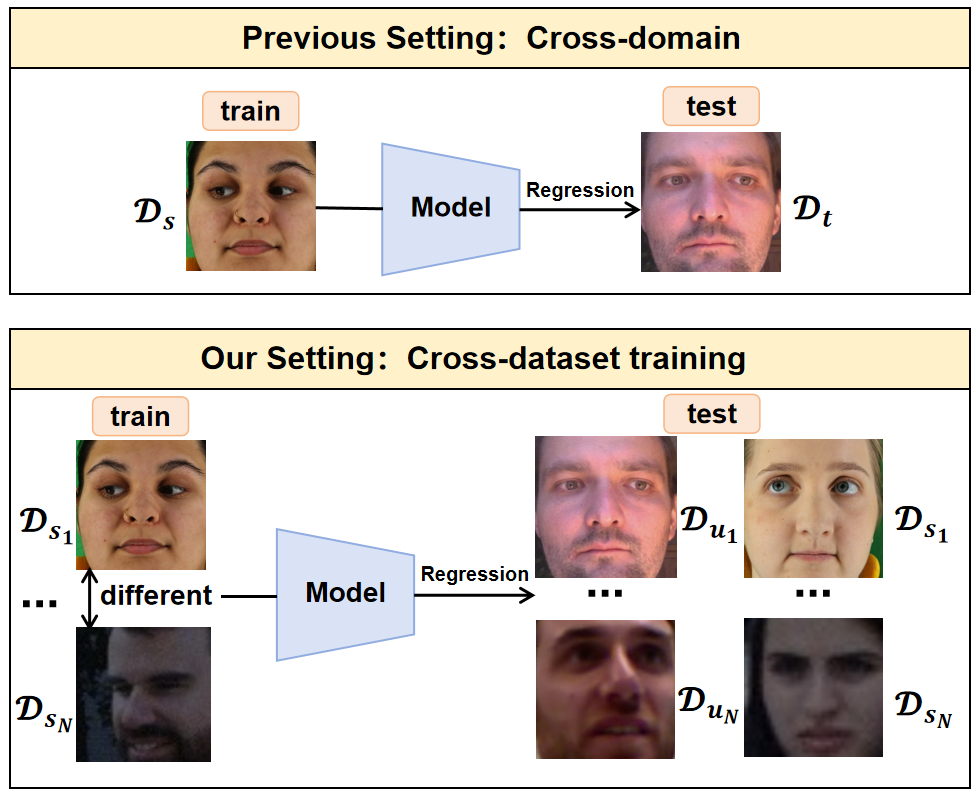}
  \caption{Comparison between traditional \textit{Cross-Domain} scenario and our \textit{Cross-Dataset Training} scenario. Here, \( \mathcal{D}_s \) represents the source domain, \( \mathcal{D}_t \) represents the target domain. \textit{Cross-Dataset Training} involves training on multiple source domains with significant distribution discrepancies, aiming for consistent performance across source domains and target domains.}
  \label{intro_cross_dataset}
  \vspace{-10pt}
\end{figure}

The challenge of gaze estimation primarily lies in the complexity and diversity of real-world environments. In fact, existing datasets~\cite{funes2014eyediap, zhang2017mpiigaze, zhang2020eth, kellnhofer2019gaze360} are collected in different environments, resulting in notable variations in data distribution among them. To enhance the model's generalization ability across multiple scenarios, current researches~\cite{wang2022contrastive, guo2020domain, liu2021generalizing, Bao_2022_CVPR, cheng2022puregaze, Cai_2023_CVPR, liu2024pnp, bao2024feature} focus on studying cross-domain scenarios and improving performance through single-source domain adaptation and domain generalization strategies, achieving promising progress. However, in practical applications, we often have easy access to multiple data sources, so it is natural to expect models trained on these datasets to perform well across all these source domains as well as target domains. This issue is known as cross-dataset training. Fig.~\ref{intro_cross_dataset} illustrates the differences between our setting and traditional cross-domain setting.
While cross-dataset training has been extensively studied in many fields such as object detection~\cite{yao2020cross, chen2023scaledet, yao2020cross}, semantic segmentation~\cite{shi2021multi, wang2022cross}, facial expression recognition~\cite{zeng2018facial} and trajectory prediction~\cite{gilles2022uncertainty}, it has received little attention in gaze estimation.

\begin{figure}
 \centering
 \includegraphics[width=8cm]{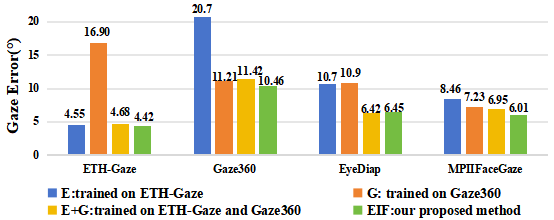}
  \caption{Test accuracy on different datasets with varied combination of training data.}
  \label{intro_pic}
  \vspace{-10pt}
\end{figure}

Preliminary experiments depicted in Fig.~\ref{intro_pic} show that simply combining data (\textit{e.g.}, ETH-Gaze+Gaze360, shorted as E+G) from multiple datasets can significantly enhance the model's performance on target domains (\textit{e.g.,} EyeDiap, MPIIFaceGaze), even outperforming carefully designed state-of-the-art (SOTA) single-source domain generalization methods~\cite{cheng2022puregaze, bao2024feature}. This improvement is due to the diverse data distributions across source domain datasets. By merging these datasets, we increase the variety of the training data, which largely enhances the model's generalization ability to target domains. Unfortunately, despite increasing the volume of training data, cross-dataset training results in diminished performance on test sets of the original source domains (ETH-Gaze, Gaze360) compared to training exclusively with their own training data. This decrease is also attributed to the significant differences in data distribution among source domains, but need to be mitigated.

To prevent the degradation of source domain performance and further enhance performance on target domains, in this paper, we propose a novel solution called Evidential Inter-intra Fusion (EIF) for cross-dataset gaze estimation. 
Acknowledging the inherent differences between the datasets, our approach involves training independent single-dataset branches for each dataset, thereby mitigating source domain performance degradation.
Additionally, building upon the findings from preliminary experiments, which suggest that merging rich source domain information benefits target domains, we devise a cross-dataset branch containing only the deep layers, where the generalizable features from each source domain branch are integrated, thereby boosting performance on target domains.

On the other hand, to combat the performance drop in the source domains, we argue that the non-stationary gazing process inherent in each dataset should be carefully considered, where non-stationary refers to our gaze being primarily influenced by head posture and eye movement with their effects varying across different gaze ranges. Such challenge is tackled by proposing to partition the data space into overlapping subspaces, with an independent local regressor assigned to each subspace. Furthermore, in order to select appropriate regressors both in within-dataset and cross-dataset branches, we design to employ evidential regressors based on the Normal and Inverse-Gamma (NIG) distribution for not only predicting gaze but also offering uncertainty estimation. Leveraging the Mixture of Normal Inverse-Gamma (MoNIG) distribution, our framework achieves both intra-evidential fusion among numerous local regressors within each dataset and inter-evidential fusion across multiple-dataset branches.

During training, to enable the model to rapidly adapt to variations in dataset combinations, the training process of our proposed EIF is divided into two stages. In stage 1, we train each single-dataset branch separately. In stage 2, we create a cross-dataset branch and jointly train all branches for only a few epochs. During inference, the predictions including both gaze and uncertainty are output, which largely enhance the reliability of our proposed framework. We thoroughly evaluate the effectiveness of our proposed EIF on established gaze estimation benchmarks.

In summary, our contribution is three-fold:

\begin{itemize}
    \item We introduce the task of cross-dataset training for the first time in the field of gaze estimation, which can significantly improve the generalization of gaze estimation while avoiding performance degradation in the source domains.
    \item We propose a novel framework, called Evidential Inter-intra Fusion (EIF) to tackle the challenge in cross-dataset gaze estimation. EIF is equipped with multiple local regressors fused by intra-evidential within each dataset and meanwhile performs inter-evidential fusion across multiple datasets, ensuring accurate gaze estimation in complex and diverse environments. Additionally, EIF is designed with a two-stage training process, enabling it to quickly adapt to variations in dataset combinations. 
    \item We conduct comprehensive experiments that demonstrate the consistent performance improvement in both source domains and target domains.
\end{itemize}

\section{Related Work}
\subsection{Gaze Estimation}
Gaze estimation methods can be divided into two main types~\cite{hansen2009eye}: model-based and appearance-based approaches. Model-based methods~\cite{alberto2014geometric, valenti2011combining} rely on the accurate reconstruction of 3D model of the eyes, thus requiring specialized equipments such as depth sensors, infrared cameras, and lighting. By contrast, appearance-based methods~\cite{cheng2021appearance} estimate gaze using images captured by a single webcam, and thus facilitate the popularity of gaze estimation. With advancements in deep learning, appearance-based methods have significantly improved the performance and gained widespread attention. However, deep learning methods heavily rely on data that can ideally reflect the distribution of real-world conditions. Yet, the actual gaze estimation scenario is diverse and complex~\cite{ghosh2023automatic}. Specifically, we provide a visual display of the different dataset collection environments and example samples in the supplementary.

In fact, 
the current gaze datasets are collected in various environments, resulting in significant differences in data distribution. For example, MPIIFaceGaze~\cite{zhang2017mpiigaze} is collected during the natural use of the laptops, with images captured using the laptop camera. EyeDiap~\cite{funes2014eyediap}, on the other hand, is collected in laboratory settings, where participants are instructed to observe given targets, and images are captured using depth cameras. Gaze360~\cite{kellnhofer2019gaze360} is collected in the wild using a 360-degree camera to simultaneously record multiple participants. Lastly, ETH-Gaze~\cite{zhang2020eth} is constructed in a laboratory green screen environment, where lighting conditions are simulated using controlled lighting, and images are captured using ultra-high-resolution cameras.

Due to variations in data distributions among datasets, traditional deep learning methods often struggle in cross-domain scenarios. Many researches focus on employing domain generalization and domain adaptation strategies to alleviate the effect of domain discrepancy. Liu \textit{et al.}~\cite{liu2021generalizing} propose outlier-guided collaborative adaptation. Wang \textit{et al.}~\cite{wang2022contrastive} introduce regression contrastive learning techniques. Bao \textit{et al.}~\cite{Bao_2022_CVPR} propose a rotation consistency strategy. Cheng \textit{et al.}~\cite{cheng2022puregaze} utilize adversarial reconstruction techniques to purify gaze features. Cai \textit{et al.}~\cite{Cai_2023_CVPR} attempt to reducing sample and model uncertainty. Bao \textit{et al.}~\cite{bao2024feature} propose constructing Physics-Consistent Features (PCF) to incorporate physical definition. These methods enhance the performance of the target domain through carefully designed strategies, but they may do so at the cost of sacrificing source domain performance~\cite{bao2024feature}.

Unlike the methods mentioned above, our work introduces a novel practical scenario called cross-dataset training. In this scenario, we have access to data from multiple source domains collected in different environments with distribution variations during training, which significantly improves target domain performance without sacrificing performance of source domain.

\subsection{Cross-dataset Training}

In many computer vision tasks, researches on cross-dataset training have been well-explored, but they face different challenges compared to ``cross-dataset gaze estimation''. For instance, in object detection, the main challenge lies in the disparate definitions and granularity of classes across different datasets. Therefore, the focus of related works~\cite{yao2020cross, chen2023scaledet, shi2021multi} is on integrating diverse datasets into a unified taxonomy. In the field of facial expression recognition, a primary challenge in cross-dataset training stems from the inconsistent annotations among datasets. To address this issue, \textit{Zeng et al.}~\cite{zeng2018facial} learn a transition matrix for modeling the relationship between the latent ground truth and the dataset annotation. 

Unlike the tasks mentioned above, the challenge need to be solved in cross-dataset gaze estimation is the inherent distribution shift across different datasets. Similar challenges have also been identified in semantic segmentation for autonomous driving by Wang \textit{et al.}~\cite{wang2022cross}. They propose a solution based on the characteristics of their task, which involves training independent batch normalization (BN) layers for each dataset while sharing their convolutional layers. However, it requires prior knowledge of which source domain the test sample belongs to for selecting the corresponding BN layer during testing. In contrast, our objective is to enable testing in both source domains and target domains for gaze estimation task without needing to know which specific source domain dataset.
\section{Methods}

\begin{figure*}
 \centering
 \includegraphics[width=18cm]{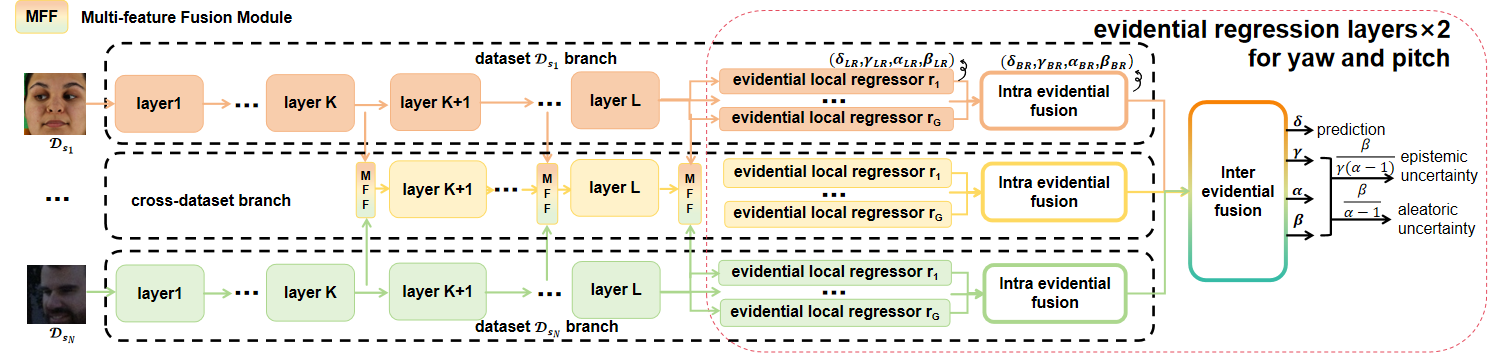}
  \caption{Illustration of the proposed Evidential Inter-intra Fusion (EIF) framework. The model consists of multiple single-dataset branches and a cross-dataset branch, with each branch comprising multiple local regressors. We utilize the intra-evidential fusion module to perform local regressor fusion within each branch and the inter-evidential fusion module to perform fusion across multiple branches. These evidence fusion modules are formulated without any learnable parameters. The output includes both predictions and uncertainty. Due to gaze estimation involving yaw and pitch components, there are two identical regression layers with the same model structure.}
  \label{model}
  \vspace{-10pt}
\end{figure*}

In this section, we elaborate  the proposed Evidential Inter-intra Fusion (EIF) framework. As depicted in Fig.~\ref{model}, EIF model incorporates multiple {single-dataset branches} {(Sec.~\ref{sec:single})}, each dedicated to a distinct source domain dataset. Additionally, to enhance performance on target domains, we introduce a cross-dataset branch {(Sec.~\ref{sec:crossdataset})} that aggregates features from all source domain branches. To address the performance degradation of source domain caused by non-stationary nature of the gazing process, we draw inspiration from similar challenges addressed in other tasks such as age estimation~\cite{li2019bridgenet}, action quality assessment~\cite{yu2021group}, equipping each branch with multiple {local regressors} tailored to specific data subspaces. Subsequently, we employ {intra-evidential fusion module} {(Sec.~\ref{sec:fuse})} to fuse local regressors within each branch and {inter-evidential fusion module} {(Sec.~\ref{sec:fuse})} to fuse multiple branches. 

\subsection{Single-dataset Branch}
\label{sec:single}
Due to inherent distributional shifts among datasets, each single-dataset branch is tailored to a specific source domain dataset. It includes several layers for feature extraction and two regression layers, \( R_1 \) and \( R_2 \), designed to estimate the two components of gaze: yaw (\( Y_1 \)) and pitch (\( Y_2 \)). Since these layers have the same structure, differing only in the gaze components they estimate, we use \( R \) for the regression layers and \( Y \) for the gaze components to introduce the model structure conveniently in the following description.

Recognizing the non-stationary nature of gaze behavior might lead to the performance drop when combining multiple datasets, we design the regression layer \( R \) to include multiple local regressors, denoted as \( R = \{ r_1, r_2, \ldots, r_G \} \). The outputs of these local regressors are centered around specific gaze labels \( Y \), thereby forming a collective of experts, each specializing in a designated gaze region and corresponding regression scenario.

Moving forward, the method of partitioning gaze label intervals \( Y \) is crucial. Given the non-uniform distribution of gaze labels, a straightforward division may lead to imbalanced data in each gaze subspace. To address this, we introduce a density-based partitioning approach. First, we sort the gaze labels \( Y = \{y^1, y^2, \ldots, y^S\} \) in ascending order to produce \( Y^* = \{y^{1^*}, y^{2^*}, \ldots, y^{S^*}\} \). We then sample indices at regular intervals to locate the mean position for each gaze group:
\begin{equation}
\zeta_{\text{center}}^g = Y^*\left(\left\lfloor S \times \frac{(g + 0.5)}{G} \right\rfloor\right),
\end{equation}
where \( \zeta_{\text{center}}^g \) is the center of the \( g \)-th gaze group, \( Y^*(s) \) denotes the \( s \)-th element in \( Y^* \), \( S \) is the total sample count, \( g \) is the gaze group index, and \( G \) represents the total number of gaze groups.

Additionally, we employ a dense overlap strategy among gaze groups, enabling adjacent groups to have overlapping intervals. This strategy ensures that data within overlapping regions contribute to training multiple regressors, thereby enhancing model robustness and accuracy across varying gaze behaviors. The degree of overlap between them is determined by a dense overlap coefficient denoted by \( \alpha \). Formally, the boundaries and length for each group are defined as follows:
\begin{equation} \label{eq:gaze_group_interval}
\begin{split}
\zeta_{\text{left}}^g & = Y^*\left(\left\lfloor S \times \frac{(g+0.5-0.5 \times \alpha)}{G} \right\rfloor\right), \\
\zeta_{\text{right}}^g & = Y^*\left(\left\lfloor S \times \frac{(g+0.5+0.5 \times \alpha)}{G} \right\rfloor\right), \\
\mathcal{I}^g & = (\zeta_{\text{left}}^g, \zeta_{\text{right}}^g), \quad L^g = \zeta_{\text{right}}^g - \zeta_{\text{left}}^g,
\end{split}
\end{equation}
where \( \mathcal{I}^g \) represents the boundaries of the \( g \)-th gaze group, and \( L^g \) denotes its length. We set \( \alpha > 1 \) to ensure that each gaze group has an overlap interval. By partitioning the gaze groups in this manner, data with gaze label \( y^{s} \) that fall within \( (\zeta_{\text{left}}^g , \zeta_{\text{right}}^g) \) are utilized for training the branch \( r_g \). 

In this setup, each regressor \( r_g \) is responsible for a specific gaze group and outputs an offset \( \Delta^g \). The final prediction is derived by combining this offset with the center \( \zeta_{\text{center}}^g \) and the interval length \( L^g \), which results in:

\begin{equation} \label{eq:gaze_group_interval}
\delta_{LR_g} = \Delta^g \cdot \frac{L^g}{2} + \zeta_{\text{center}}^g.
\end{equation}

This offset-based approach allows each regressor to capture variations within its designated interval, thus enhancing the model's adaptability across diverse gaze behaviors.

\subsection{Cross-dataset Branch}
\label{sec:crossdataset}
To improve performance in target domains, preliminary experiments have demonstrated that utilizing data from various source domains can significantly boost the model's effectiveness in such scenarios. This insight leads to the creation of the cross-dataset branch, a key component designed to combine information from different data sources. The cross-dataset branch is equipped with multiple Multi-Feature Fusion (MFF) modules and regression layers, whose architecture mostly mirrors the higher layers of the single-dataset branch. 

Formally, let \( f_n^l \) denote the \( l \)-th layer feature of the \( n \)-th source dataset branch. Assuming the feature fusion from multiple source datasets begins at the \( K \)-th layer, the fusion process 
of the MFF module can be represented as:
\begin{equation}
f_{\text{cross}}^K = \text{MFF}_k(f_{1}^K, f_{2}^K, \ldots, f_{S_N}^K),
\end{equation}
where \( S_N \) is the total number of source datasets, and \( f_{\text{cross}}^{K} \) indicates the features output from \( K \)-th MFF layer.

After that, in the following MFF modules, we fuse features from all source dataset branches as well as the features output from the previous MFF module. This fusion process can be described as:
\begin{equation}
f_{\text{cross}}^{k+1} = \text{MFF}_{k}(f_{1}^{k}, f_{2}^{k}, \ldots, f_{S_N}^{k}, f_{\text{cross}}^{k}),
\end{equation}
where \( k \) ranges from \( K \) to \( L-1 \), and \( L \) denotes the total number of layers for feature extraction in the single-dataset branch.

The inputs of MFF module come from different branches at the same layer, so their features are spatially aligned. But they may not be aligned in channels, hence, we first independently apply multiple point-wise convolutions to align features in channels. Subsequently, we use the Attention Feature Fusion (AFF) module proposed by Dai \textit{et al.}~\cite{dai21aff} to fuse features pairwise (refer to~\cite{dai21aff} for details). The computation of \( k \)-th MFF module is expressed as:
\begin{equation}
\text{MFF}_{k} = \text{AFF}(...,\text{AFF}(\text{conv}(\cdot), \text{conv}(\cdot))).
\end{equation}

\subsection{Evidential Fusion Module}
\label{sec:fuse}
In this section, we introduce both an intra-evidential fusion module for fusing multiple local regressors within each dataset and an inter-evidential fusion module for fusing multiple dataset branches. The evidential fusion module is based on evidence regression learning \cite{amini2020deep}. From its perspective, the gaze component \( y \) is sampled from a normal distribution, with the mean and variance sampled from normal and Inverse-Gamma (NIG) distribution, respectively. Therefore, \( y \) can be seen as indirectly sampled from an NIG distribution with parameters \(\boldsymbol{\tau}=(\delta, \gamma, \alpha, \beta)\), where \(\delta \in \mathbb{R}\), \(\gamma>0\), \(\alpha>1\), and \(\beta>0\). This can be expressed as:
\begin{equation}
y \sim \mathcal{N}(\mu, \sigma^2), \ \mu \sim \mathcal{N}(\delta, \sigma^2\gamma^{-1}), \ \sigma^2 \sim \Gamma^{-1}(\alpha, \beta).
\end{equation}
where \(\Gamma(\cdot)\) is a gamma function.  All of our local regressors adopt the evidential approach which is implemented by a fully connected layer that outputs four numbers corresponding to \(\delta\), \(\gamma\), \(\alpha\), and \(\beta\). During training, the evidence learning loss \( \mathcal{L}_\text{evidence} \) consists of two essential components: a negative log-likelihood \( \mathcal{L}_\text{NLL} \) and a regularization term \( \mathcal{L}_\text{R} \). The negative log-likelihood \( \mathcal{L}_\text{NLL} \) is defined as:
\begin{align}
\begin{split}
\mathcal{L}_\text{NLL}(\delta, \gamma, \alpha, \beta) &= \frac{1}{2}\log\left(\frac{\pi}{\gamma}\right) - \alpha\log\left(\Omega\right) \\
&\quad+ \left(\alpha+\frac{1}{2}\right)\log\left((y-\delta)^2\gamma+\Omega\right) \\
&\quad+ \log\left(\frac{\Gamma(\alpha)}{\Gamma\left(\alpha+\frac{1}{2}\right)}\right),
\end{split}
\end{align}
where \( \Omega = 2\beta(1+\gamma) \). The regularization term \( \mathcal{L}_\text{R} \) is introduced to penalize incorrect evidence predictions and is given by:
\begin{equation}
\mathcal{L}_\text{R}(\delta, \gamma, \alpha, \beta, y) = |y-\delta|\cdot (2\gamma+\alpha).
\end{equation}
The total loss, denoted as \( \mathcal{L}_\text{total} \), is obtained by combining the negative log-likelihood and regularization term:
\begin{equation}
\begin{split}
\mathcal{L}_\text{evidence}(\delta, \gamma, \alpha, \beta, y) =& \mathcal{L}_\text{NLL}(\delta, \gamma, \alpha, \beta)\\
&+ \lambda \mathcal{L}_\text{R}(\delta, \gamma, \alpha, \beta, y),
\end{split}
\end{equation}
where the coefficient \( \lambda > 0 \) balances the contributions of the two loss terms.

Building upon evidence regression learning, the evidential fusion module employs the Mixture of normal-inverse gamma distribution (MoNIG)~\cite{ma2021trustworthy} for fusion. The resulting distribution remains NIG, represented as:
\begin{align}
\begin{split}
\text{MoNIG}(\delta, \gamma, \alpha, \beta) =& \text{NIG}(\delta_1, \gamma_1, \alpha_1, \beta_1) \oplus \\
& \text{NIG}(\delta_2, \gamma_2, \alpha_2, \beta_2) \oplus \\
&\cdots \oplus \text{NIG}(\delta_M, \gamma_M, \alpha_M, \beta_M),
\end{split}
\end{align}
where \( M \) is the number of NIG distributions being fused, and the parameters are computed as:
\begin{align}
\begin{split}
\label{equation:MoNIG}
\delta &= \frac{\sum_{i=1}^{M} \gamma_i \delta_i}{\sum_{i=1}^{M} \gamma_i}, \quad \gamma =  \sum_{i=1}^{M} \gamma_i, \quad
\alpha = \sum_{i=1}^{M} \alpha_i + \frac{1}{M},  \\ 
\beta &= \sum_{i=1}^{M} \beta_i + \frac{1}{M} \sum_{i=1}^{M} \gamma_i (\delta_i - \delta)^2.
\end{split}
\end{align}

Given the outputs of the \(g\) local regressors in the \(n\)-th branch as \( \text{NIG}_{\text{LR}_g}^n(\delta_{\text{LR}_g}^n, \gamma_{\text{LR}_g}^n, \alpha_{\text{LR}_g}^n, \beta_{\text{LR}_g}^n) \), the output of the \(n\)-th branch after the intra-evidential fusion module, \( \text{NIG}_\text{BR}^n \), can be represented as:
\begin{align}  \label{eq:intra_MoNIG}
\begin{split}
\text{NIG}_{\text{BR}}^n(\delta_{\text{BR}}^n, \gamma_\text{BR}^n, \alpha_{\text{BR}}^n, \beta_\text{BR}^n) =& \text{NIG}_{\text{LR}_1}^n \oplus \cdots \
\\
& \oplus \text{NIG}_{\text{LR}_G}^n.
\end{split}
\end{align}

Similarly, the output of each branch are further fused by the inter-evidential fusion module, and the final output, \( \text{NIG}(\delta, \gamma, \alpha, \beta) \), can be represented as:
\begin{align} \label{eq:inter_MoNIG}
\begin{split}
\text{NIG}(\delta, \gamma, \alpha, \beta) =& \text{NIG}_\text{BR}^1 \oplus \ldots \\
& \oplus \text{NIG}_\text{BR}^{S_N} \oplus \text{NIG}_\text{BR}^{cross},
\end{split}
\end{align}
where \( \text{NIG}_\text{BR}^{\text{cross}} \) corresponds to the output of the cross-dataset branch. Then, we can compute the prediction gaze, aleatoric, and epistemic uncertainties based on \( \text{NIG}(\delta, \gamma, \alpha, \beta) \) as follows:
\begin{align}
    \label{equation:evidence_regression_learning}
    \underbrace{\mathbb{E}[\mu]=\delta}_{\text{gaze}}, \quad \underbrace{\mathbb{E}[\sigma^2]=\tfrac{\beta}{\alpha-1}}_{\text{aleatoric}}, \quad \underbrace{\text{Var}[\mu]=\tfrac{\beta}{\gamma(\alpha-1)}}_{\text{epistemic}}.
\end{align}

\subsection{Two-stage Training Process}

To enable the model to rapidly adapt to variations in the number of datasets, we devise a two-stage training strategy. 
In stage 1, we train each source domain branch separately. 
For each dataset branch, the corresponding source domain data is divided into \(G\) gaze groups \(\mathcal{I}^g\) based on Eq.~\ref{eq:gaze_group_interval}. The output of \(G\) local regressors for the \(s\)-th sample of the source domain data is denoted as \(\delta_{\text{LR}_g}^s, \gamma_{\text{LR}_g}^s, \alpha_{\text{LR}_g}^s, \beta_{\text{LR}_g}^s\). We only compute \(\mathcal{L}_{\text{local}}^g\) within the \(g\)-th gaze group data subspace corresponding to the \(g\)-th local regressor, as follows:
\begin{equation}
\mathcal{L}_{\text{local}}^g = \sum_{y^s \in \mathcal{I}^g} \mathcal{L}_{\text{evidence}}(\delta_{\text{LR}_g}^s, \gamma_{\text{LR}_g}^s, \alpha_{\text{LR}_g}^s, \beta_{\text{LR}_g}^s, y^s).
\end{equation}
Here, \(y^s\) represents the gaze label of the \(s\)-th sample.
Then, we compute the output of this dataset branch after intra-evidential fusion according to Eq.~\ref{eq:intra_MoNIG}, denoted as \(\delta_\text{BR}^s, \gamma_\text{BR}^s, \alpha_\text{BR}^s, \beta_\text{BR}^s\).
We compute the loss \(\mathcal{L}_{\text{global}}\) for it across the entire source domain dataset as follows:
\begin{equation}
\mathcal{L}_{\text{global}} = \sum \mathcal{L}_{\text{evidence}}(\delta_\text{BR}^s, \gamma_\text{BR}^s, \alpha_\text{BR}^s, \beta_\text{BR}^s, y^s)
\end{equation}

Therefore, during the first stage of training, the loss \(\mathcal{L}_{\text{branch}}\) for each single-dataset branch is as follows:
\begin{equation} \label{eq:loss_branch}
\mathcal{L}_{\text{branch}} = \sum_{g=1}^{G} \mathcal{L}_{\text{local}}^g + \mathcal{L}_{\text{global}}.
\end{equation}

In Stage 2, we create a cross-dataset branch and jointly train all branches for a few epochs. We merge multiple source domain datasets and use Eq.~\ref{eq:gaze_group_interval} to obtain \(G\) gaze groups, each characterized by center \(\zeta_{\text{center}}^g\) and interval length \(L^g\). The cross-dataset branch loss \(\mathcal{L}_{\text{cross}}\) is computed according to Eq.~\ref{eq:loss_branch}, but utilizing all source datasets. Given the outputs of each branch, we fuse them using Eq.~\ref{eq:inter_MoNIG} to obtain \(\delta, \gamma, \alpha, \beta\) and compute the joint loss \(\mathcal{L}_{\text{joint}}\) across all source domain datasets:
\begin{equation}
\mathcal{L}_{\text{joint}} = \sum_{\mathcal{D}_{S_1}...\mathcal{D}_{S_N}} \mathcal{L}_{\text{evidence}} (\delta, \gamma, \alpha, \beta, y).
\end{equation}

Thus, the total loss for Stage 2 is:
\begin{equation}
\mathcal{L}_{\text{stage2}} = \mathcal{L}_{\text{cross}} + \mathcal{L}_{\text{joint}}.
\end{equation}

In this approach, each single-dataset branch is trained independently in Stage 1, allowing for model reuse. Thus, various model combinations can be achieved by training only in Stage 2. Furthermore, to promote effective learning in the cross-dataset branch, each source domain dataset is split in a specific ratio across the two stages, ensuring that the data used in joint training remains distinct from that in Stage 1, thereby reducing the risk of overfitting and enhancing cross-dataset branch learning.

\captionsetup[table]{skip=5pt} 
\section{Experiment}
\subsection{Cross-dataset Setting}
The experiments utilize four widely used gaze estimation datasets, EyeDiap~\cite{funes2014eyediap}, MPIIFaceGaze~\cite{zhang2017mpiigaze}, Gaze360~\cite{kellnhofer2019gaze360}, and ETH-XGaze~\cite{zhang2020eth}. We pre-process the data following the technique in~\cite{cheng2021appearance}. To facilitate comparisons, we reference the setup of cross-domain scenarios~\cite{xu2023learning, Cai_2023_CVPR} in our cross-dataset setting, \textit{i.e.,} we choose ETH-XGaze and Gaze360, known for their wider gaze distributions and larger data volumes, as the source domains, and MPIIGaze and EyeDiap as the target datasets. Detailed information about the datasets can be found in the Supplementary Material.





\subsection{Implementation Details}
We use ResNet-18 backbone following previous studies \cite{cheng2022puregaze, Cai_2023_CVPR, bao2024feature}. All images are resized to 224 × 224, and we use the Adam optimizer with a learning rate of \(10^{-4}\).
In Stage 1, we train on each source domain dataset for 100k batches. Specifically, for the Gaze360 single-dataset branch, we use the ETH-Gaze’s backbone for pre-training, similar to the previous work~\cite{cheng2022gaze}.
In stage 2, we perform joint training for only 5k batches using all source domain datasets. Considering the discrepancy in the quantity of data among source domain datasets, during the joint training in stage 2, instead of directly mixing multiple datasets for sampling, we adopt a balanced mixing approach, where the datasets with fewer samples are oversampling to balance the data across all source domain datasets. 
Notably, our two-stage learning strategy allows for quick adaptation. Once the single-dataset branches are trained, Stage 2 requires only a few epochs to adjust. In our experiments, Stage 2 takes 40 minutes on two NVIDIA TITAN XP GPUs, while the baseline takes 5 hours.
By default, we set the number of local regressors \( G \) to 8, the overlap coefficient \( \alpha \) to 2.0, the evidence learning regularization loss \( \lambda \) to 0.01, the dataset split ratio in stages 1 and 2 to 4:1, and the starting MFF layer \( K \) in the cross-dataset branch to 3.

\begin{table*}[t]
\centering
\small
\caption{Comparison of gaze estimation performance (angle error, °) in domain generalization. The best, second best, and third best results are denoted as \BEST{red}, \SBEST{blue}, and \TBEST{green}, respectively. † indicates the reproduced cross-domain gaze estimation results. Since the AGG code has not yet been open-sourced and certain results were not disclosed in their paper, we have left some sections of the tables unfilled.}
\label{tab:cross_dataset_training}
\resizebox{0.95\textwidth}{!}{  
\begin{tabular*}{\textwidth}{@{\extracolsep{\fill}}ccccccccc@{}}
\toprule
\multicolumn{3}{c}{\textbf{Method}} & \multicolumn{2}{c}{\textbf{Source Domain}} & \multicolumn{2}{c}{\textbf{Target Domain}} & \multicolumn{2}{c}{\textbf{Average}} \\
\cmidrule(lr){1-3} \cmidrule(lr){4-5} \cmidrule(lr){6-7} \cmidrule(lr){8-9} 
\textbf{Type} & \textbf{Name} & \textbf{Train Data} & \textbf{ETH} & \textbf{G360} & \textbf{EyeDiap} & \textbf{MPII} & \textbf{Target} & \textbf{All} \\
\midrule
\multirow{2}{*}{\shortstack{Single-dataset\\Baseline}} 
& Baseline & ETH & \SBEST{4.55} & 20.70 & 10.70 & 8.46 & 9.58 & 11.10 \\
& Baseline & G360 & 16.90 & 11.21 & 10.90 & 7.23 & 9.07 & 11.56 \\
\midrule
\multirow{2}{*}{\shortstack{Cross-dataset\\Baseline}} 
& Simple Mixing & ETH+G360 & \TBEST{4.68} & 11.42 & \SBEST{6.42} & 6.95 & 6.69 & \TBEST{7.37} \\
& Balanced Mixing & ETH+G360 & 4.90 & \SBEST{10.76} & \BEST{6.40} & \SBEST{6.50} & \SBEST{6.45} & \SBEST{7.14} \\
\midrule
\multirow{2}{*}{\shortstack{Multi-source\\Domain Generalization}}
& GMDG~\cite{tan2024rethinking} & ETH+G360 & 4.76 & \TBEST{10.95} & 7.12 & 7.07 & 7.10 & 7.48 \\
& BNE~\cite{segu2023batch} & ETH+G360 & 5.44 & 11.06 & 6.73 & \TBEST{6.51} & \TBEST{6.62} & 7.44 \\
\midrule
\multirow{4}{*}{\shortstack{Cross-domain\\Gaze Estimation}} 
& PureGaze~\cite{cheng2022puregaze} & ETH & 4.77† & 21.67† & 7.48 & 7.08 & 7.28 & 10.25 \\
& PureGaze~\cite{cheng2022puregaze} & G360 & 17.11† & 11.21† & 9.32 & 9.28 & 9.30 & 11.73 \\
& AGG~\cite{bao2024feature} & ETH & 5.56 & - & 7.07 & 7.10 & 7.09 & - \\
& AGG~\cite{bao2024feature} & G360 & - & 13.03 & 7.93 & 7.87 & 7.90 & - \\
\midrule
\multirow{1}{*}{Our Method} 
& EIF & ETH+G360 & \BEST{4.42} & \BEST{10.46} & \TBEST{6.45} & \BEST{6.01} & \BEST{6.23} & \BEST{6.84} \\
\bottomrule
\end{tabular*}
}
\vspace{-10pt}
\end{table*}

\subsection{Domain Generalization Results}
We compared our method with four types of approaches:

\noindent\textbf{Single-dataset baselines} are trained on a single source domain using the standard L1 loss function for gaze estimation, while other network settings remain consistent with those of our proposed method.

\noindent\textbf{Cross-dataset baselines} are trained on multi-source using two different mixing strategies: \textit{Simple Mixing}, where the datasets are directly concatenated, and \textit{Balanced Mixing}, which, similar to our Stage 2 method, uses balanced sampling. The network and training settings are kept identical to those in the single-dataset baseline.

\noindent\textbf{Multi-source domain generalization methods} aim to explore combining multiple source datasets to enhance performance in the target domain, making them relevant to our scenario. We implement two representative regression methods: GMDG~\cite{tan2024rethinking} and BNE~\cite{segu2023batch}, with training settings consistent with the balanced mixing cross-dataset baseline. Model details are provided in the supplementary.

\noindent\textbf{Cross-domain gaze estimation} are trained on a single source domain using carefully designed algorithms. We report two state-of-the-art approaches~\cite{cheng2022puregaze, bao2024feature} for comparison. 

From Table \ref{tab:cross_dataset_training}, it can be observed that the cross-dataset baseline  achieves comparable performance with the carefully designed SOTA single-domain generalization methods on the target dataset. However, there is a slight error increase in the source domains. Multi-source domain generalization fails to bring improvements and may even slightly degrade performance. In contrast, our proposed method EIF reduces gaze errors in both the source domains (ETH-Gaze by 0.48° and Gaze360 by 0.30°) and the target domains (by 0.22°) compared to the best cross-dataset baseline (balanced mixing), ultimately surpassing the best  cross-domain generalization method AGG~\cite{bao2024feature} (ETH-Gaze) by a large improvement (1.14° in source domain and 0.86° in target domains).

\subsection{Domain Adaptation Results} 
Our previous experiments demonstrate the effectiveness of our method both in the source domain and in generalizing to target domains. To further validate the efficacy of our EIF framework, as done in previous domain generalization works~\cite{cheng2022puregaze, bao2023pcfgaze}, we conduct domain adaptation experiments. 

Similar to the previous domain adaptation setting~\cite{bao2023pcfgaze}, we fine-tune the model using 100 images from the target domain over 100 batches. For our EIF, we use these data to constrain the evidence loss (Eq.~\ref{equation:evidence_regression_learning}), both for all branches and their joint output after inter-fusion. To mitigate randomness from selecting the training set, we repeat the experiment 5 times and report the average results.

As shown in Table \ref{table: domain adaption}, our EIF framework significantly improves performance, reducing the error by 0.83° compared to the second best method, PCFGaze (ETH-Gaze).

\begin{table}[h]
\centering
\small
\tabcolsep=1.5mm
\caption{Comparison with SOTA methods in domain adaptation on 100 labeled samples. Results are shown in angle error (°).}
\label{table: domain adaption}
\begin{tabular} {lccc}
\toprule
\multirow{2}{*}{Methods} & \multicolumn{2}{c}{Target Dataset} & \multirow{2}{*}{Avg}\\
\cmidrule(lr){2-3}
& MPII & EyeDiap \\
\midrule
PureGaze (ETH-Gaze)~\cite{cheng2022puregaze} & 5.30 & 6.42 & 5.86\\
PureGaze (Gaze360)~\cite{cheng2022puregaze} & 5.20 & 7.36 & 6.28\\
PCFGaze (ETH-Gaze)~\cite{bao2023pcfgaze} & 4.74 & 5.88 & 5.31\\
PCFGaze (Gaze360)~\cite{bao2023pcfgaze} & 5.41 & 6.39 & 5.90 \\
\midrule
EIF (ETH-Gaze+Gaze360) & \textbf{3.91} & \textbf{5.04} & \textbf{4.48}\\
\bottomrule
\end{tabular}
\vspace{-5pt}
\end{table}

\subsection{Ablation Study}
Our model consists of two stages: single-dataset training to train a well-performing evidence learning regression branch for each dataset, and cross-dataset joint training to fuse these branches for improving cross-dataset generalization. To verify, we conduct ablation experiments in two parts.

\textbf{Single-dataset training.} It primarily includes two components: evidential regression learning and the use of multiple local regressors with intra-evidential fusion to aggregate them.
\begin{table}[t]
\centering
\small
\caption{Ablation study of the single-dataset training  on the ETH-Gaze Dataset.}
\scalebox{0.9}{
\begin{tabular}{ccc|c}
\hline
\begin{tabular}[c]{@{}c@{}}L1 \\ Regression\end{tabular} & \begin{tabular}[c]{@{}c@{}}Evidential \\ Regression\end{tabular} & \begin{tabular}[c]{@{}c@{}}Intra-Evidential \\ Fusion\end{tabular} &  Angle Error(°)  \\ \hline
 $\checkmark$  & &  &  4.55  \\
 & $\checkmark$  &  &  4.59  \\
 & $\checkmark$  &$\checkmark$   & \textbf{4.35} \\ \hline
\end{tabular}}
\label{table: ablation_study_single_dataset}
\end{table}
As shown in Table~\ref{table: ablation_study_single_dataset}, we evaluate the components through ablation studies: 1) \textit{Evidence regression} introduces complex distributions to estimate uncertainty, slightly increasing the error by 0.04°, but providing the foundation for inter-intra evidence fusion. 2) \textit{Intra-evidential fusion} reduces angle error by 0.24° compared to evidential regression alone, surpassing the baseline method without evidence regression by 0.20°.

\begin{table}[t]
\centering
\caption{Ablation study for cross-dataset joint training. It showcases the average angle errors (°) of the models trained on multiple source domains (ETH+G360) across the source domain test sets (ETH and G360) and the target domains (MPII and EyeDiap).}
 \vspace{-3pt}
\scalebox{0.85}{
\begin{tabular}{ccc|cc}
\hline
\begin{tabular}[c]{@{}c@{}}
Average \\ Fusion\end{tabular} & \begin{tabular}[c]{@{}c@{}}Inter-Evidential \\ Fusion\end{tabular} & \begin{tabular}[c]{@{}c@{}}Cross-Dataset \\ Branch \end{tabular} &  Source & target  \\ \hline
 $\checkmark$  & &  &  11.32  & 6.86\\
 & $\checkmark$  &  &  7.47  & 6.67 \\
 & $\checkmark$  &$\checkmark$   & \textbf{7.44}  & \textbf{6.23} \\ \hline
\end{tabular}}
\label{table: ablation_cross_dataset_jointly}
 \vspace{-10pt}
\end{table}

\textbf{Cross-dataset joint training.} To jointly train multiple single-dataset branches, we use the inter-evidential fusion module to aggregate predictions and design a cross-dataset branch to integrate features. As shown in Table~\ref{table: ablation_cross_dataset_jointly}, ablation studies show that: 1) \textit{Inter-evidential fusion} reduces errors by 3.85° and 0.19° in the source and target domains, respectively, compared to average fusion, which performs well in the target domain but worse in the source domain; 2) \textit{Cross-dataset branch} further reduces errors by 0.03° and 0.44° in the source and target domains, respectively. These results demonstrate that the cross-dataset branch effectively aggregates features from each source domain, particularly enhancing generalization to the target domain.

\subsection{Parameter Sensitivity Analysis}
\begin{figure}
 \centering
 \includegraphics[width=8.5cm]{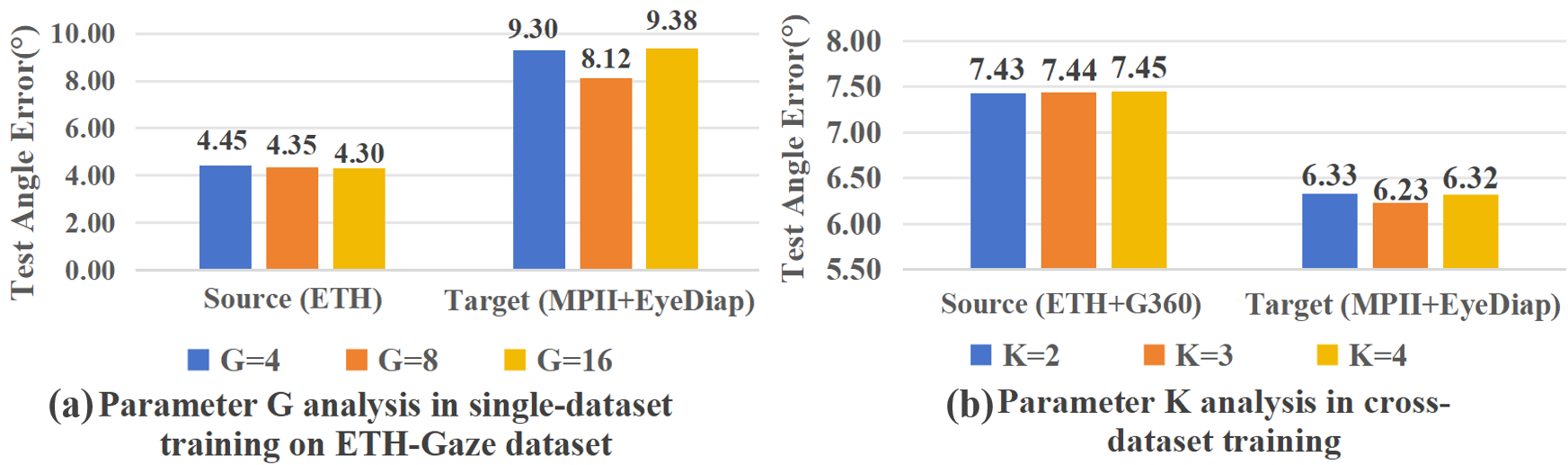}
  \vspace{-15pt}
  \caption{Parameter Sensitivity Analysis.}
  \label{image:parameter}
  \vspace{-5pt}
\end{figure}

We consider two critical hyperparameters: 1) \textit{The number of local regressors \(G\) in single-dataset training}: As shown in Fig.~\ref{image:parameter}(a), adjusting \(G\) within a certain range has a stable impact on performance, with \(G = 8\) yielding the best results. 2) \textit{The MFF starting layer \(K\) in cross-dataset joint training}: As shown in Fig.~\ref{image:parameter}(b), adjusting \(K\) also stabilizes performance within a certain range, with \(K = 3\) achieving the optimal performance.

\subsection{Evidence Analysis}
Evidential learning is crucial to achieve an inter-intra fusion in our framework. Besides, the final evidential regression output can also provide uncertainty estimations for robust predictions. Thus, in this section, we offer a visual analysis of the fusion process and uncertainty prediction to validate the effectiveness of evidential learning in gaze estimation.


\textbf{Inter-intra fusion visual analysis.} 
\begin{figure}
 \centering
 \includegraphics[width=8.5cm]{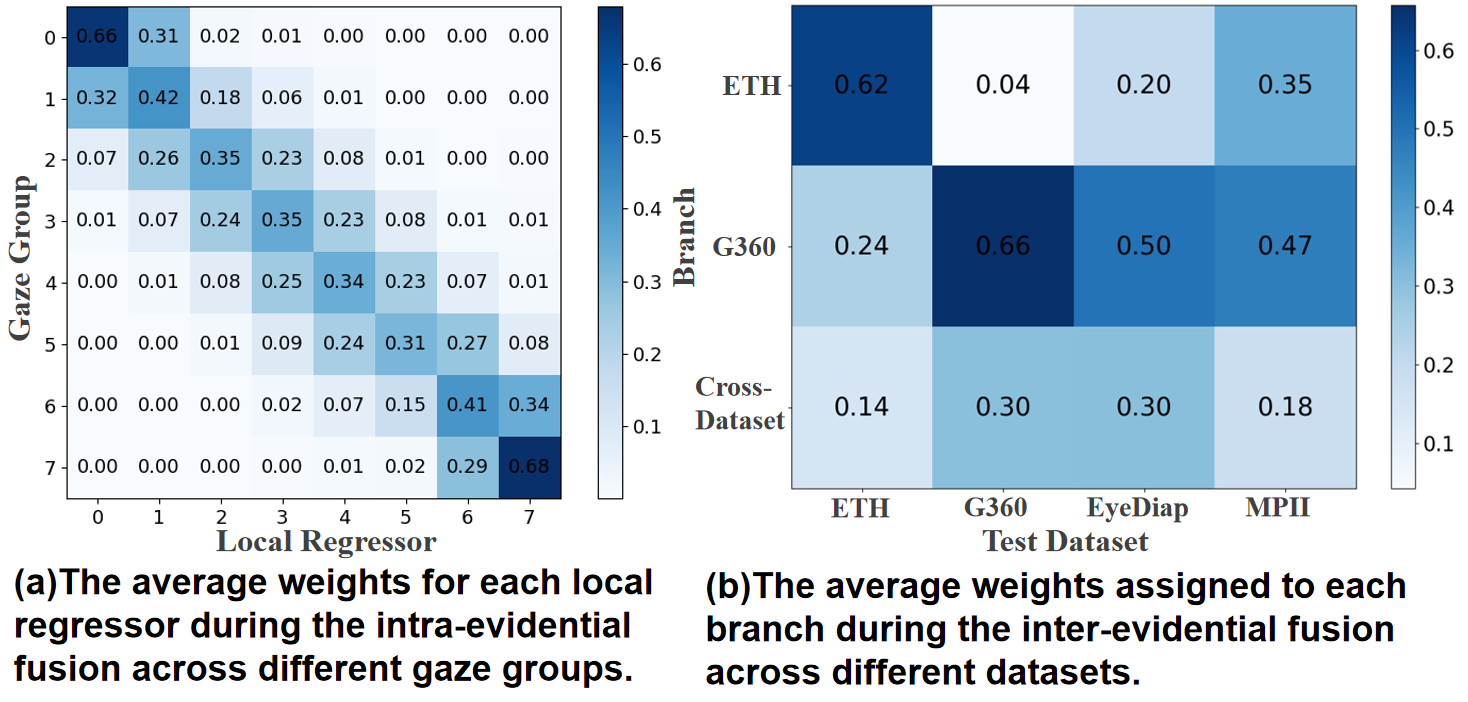}
   \vspace{-18pt}
  \caption{Visualizing Weights During Inter-Intra Fusion.}
  \label{image:inter-intra_visualize}
   \vspace{-12pt} 
\end{figure}
According to Eq.\ref{equation:MoNIG}, during the evidential fusion process, normalized \(\gamma\) is used as weights to aggregate the outputs \(\delta\). Therefore, we present the normalized \(\gamma\) values for the intra-fusion of the single-dataset branch in ETH-Gaze and the inter-fusion of EIF. In Fig.~\ref{image:inter-intra_visualize}(a), the highest values appear in the diagonal region, indicating that the intra-fusion module effectively matches samples within each gaze group to their corresponding local regressors. In Fig.~\ref{image:inter-intra_visualize}(b), source domain test sets assign the highest weights to their respective single-dataset branches, while target domains distribute the weights more evenly across branches, which aligns with our expectations.

\textbf{Uncertainty estimation analysis.} 
\begin{figure}
 \centering
 \includegraphics[width=8.5cm]{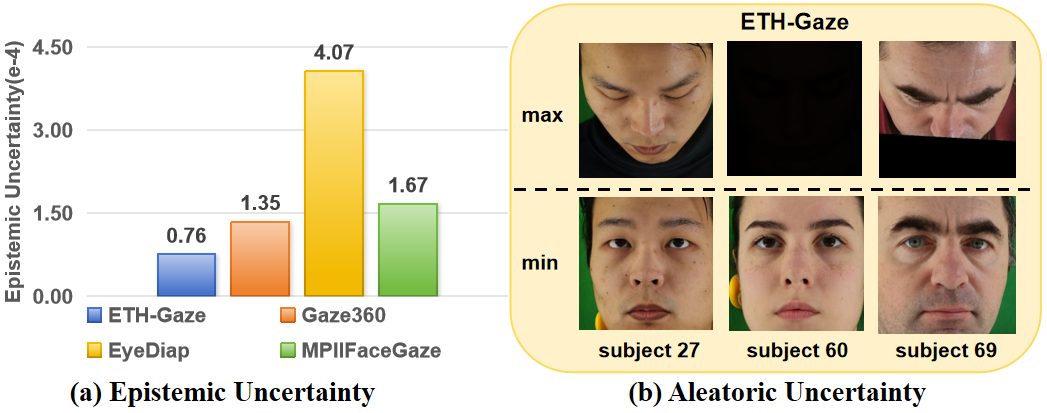}
 \vspace{-15pt}
  \caption{Visualizing Epistemic and Aleatoric Uncertainty.}
  \label{image:uncertainy}
  \vspace{-15pt}
\end{figure}
We visualize the epistemic and aleatoric uncertainty output by EIF in Fig. \ref{image:uncertainy}. To mitigate gaze range differences across datasets, epistemic uncertainty is computed only for gaze within ±20° for all datasets. In Fig.~\ref{image:uncertainy}(a), epistemic uncertainty is smaller in the source domain (ETH+G360) compared to the target domain (EyeDiap+MPII), with MPII showing lower uncertainty than EyeDiap, likely due to its closer similarity to the source domain. Fig.~\ref{image:uncertainy}(b) shows that images with higher aleatoric uncertainty indeed appear more ambiguous. More details are provided in the supplementary.

\section{Conclusion and Discussion}
Training with diverse source domain data enhances model generalization in target domains. However, simple mixing can worsen domain-specific errors due to distribution shifts. To address this, we propose a two-stage training approach. In stage 1, we use single dataset branches for each source domain. In stage 2, we combine these branches with a cross-dataset branch to aggregate features. Given the non-stationary nature of gaze behavior, we divide each source domain into subsets with local regressors. We apply evidential fusion to integrate predictions from all branches and regressors. Experiments show the effectiveness of our approach in both source and target domains.

\textbf{Limitation.} While our EIF method has demonstrated significant improvements in cross-dataset training with two commonly used, highly divergent source domains, a limitation is the lack of consideration for handling an increasing number of source domain datasets with similarities. Specifically, using a single branch shared by multiple similar domains could help save resources. To address this, we plan to explore clustering similar datasets and allowing them to share a single branch, thus improving the scalability. 

{
    \small
    \bibliographystyle{ieeenat_fullname}
    \bibliography{main}
}

\clearpage
\setcounter{page}{1}
\setcounter{section}{0}
\maketitlesupplementary

In this supplementary material, we provide further details on our proposed methods and experiments. The first section offers a detailed introduction to the dataset used in our cross-dataset gaze estimation setting, including a visual display of the different dataset collection environments, along with example samples and their label distributions. This section highlights the differences between the datasets and underscores the necessity of our work in cross-dataset gaze estimation. The second section provides an overview of the representative multi-source domain generalization methods we selected for comparison. The third section presents additional visualization results related to aleatoric uncertainty. Together, these sections provide a comprehensive illustration of our methodology.
\section{Detailed Dataset Descriptions}
In this section, we will provide a detailed introduction to the gaze estimation datasets. This explanation will help further clarify the motivation behind our proposed cross-dataset gaze estimation setting and the design of our method.

\begin{figure*}[b]
 \centering
 \includegraphics[width=14cm]{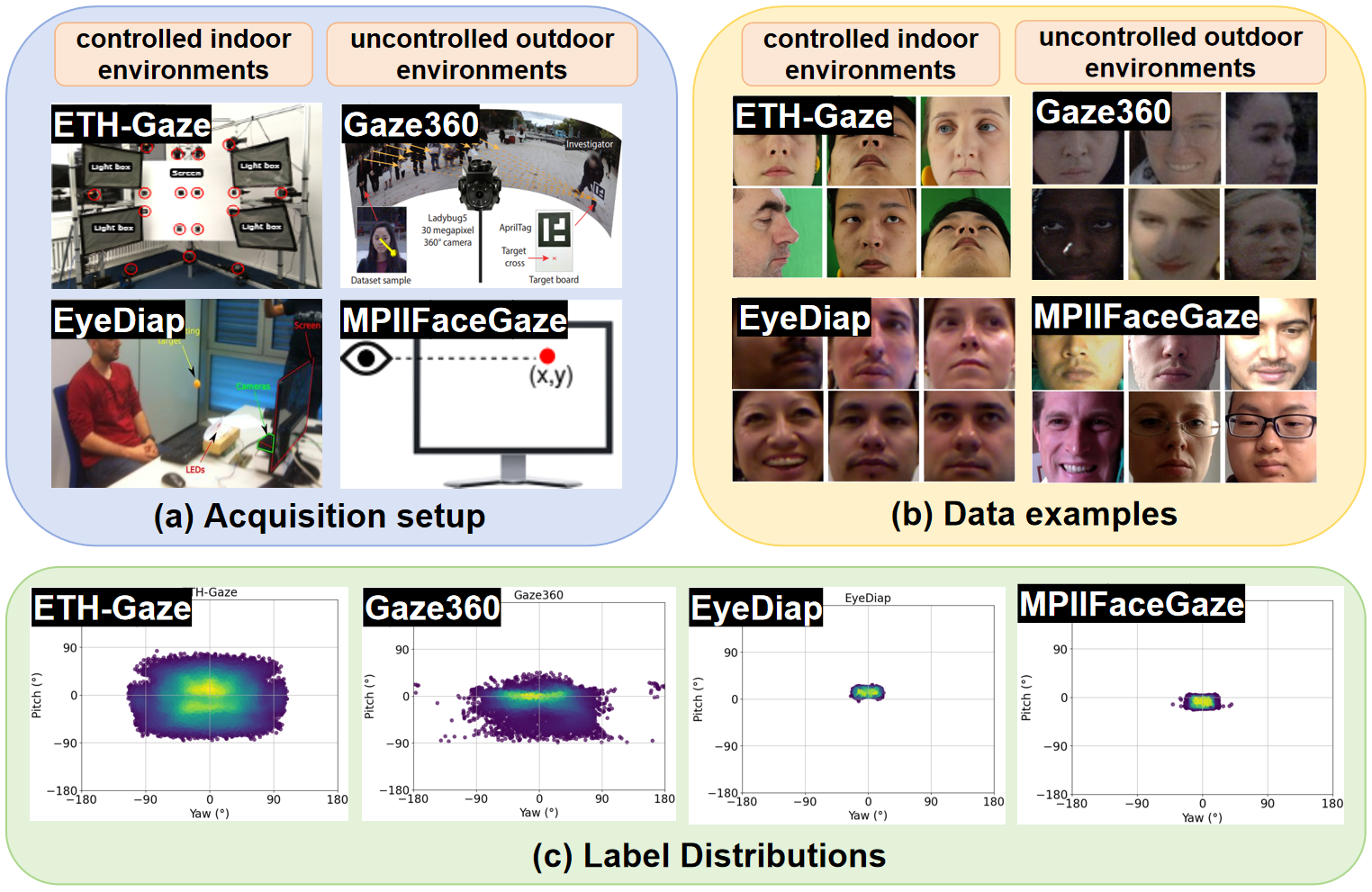}
  \caption{Visual display of the different dataset collection environments, along with data examples and their label distributions.}
  \label{image:realted_dataset}
   \vspace{-8pt} 
\end{figure*}

The experiments utilize four widely used gaze estimation datasets: EyeDiap~\cite{funes2014eyediap}, MPIIFaceGaze~\cite{zhang2017mpiigaze}, Gaze360~\cite{kellnhofer2019gaze360}, and ETH-XGaze~\cite{zhang2020eth}. For a fair comparison, the data partitioning and preprocessing techniques for these datasets are kept consistent with prior studies, as outlined in \cite{cheng2021appearance}. As illustrated in Fig~\ref{image:realted_dataset}, these datasets were collected under different setups and environments, resulting in substantial differences in the data samples and label distributions. Below, we provide a detailed description of each dataset.

\textbf{ETH-Gaze:} This dataset is captured in a laboratory environment using high-resolution cameras and consists of 756k training images from 80 subjects, exhibiting a wide range of gaze label variations. Since the ETH-Gaze dataset does not include gaze target annotations for the test set, we redefine the dataset splits following the approach in \cite{cheng2021appearance}. Specifically, we split the original training set into two subsets to use as training and testing sets, as done in previous studies \cite{cheng2023dvgaze}.

\textbf{Gaze360:} This dataset is collected in uncontrolled environments using a 360-degree camera. It includes 84k training images from 54 subjects and 16k testing images from 15 subjects, featuring a wide variety of gaze label distributions.

\textbf{MPIIFaceGaze:} This dataset is captured during natural laptop use, utilizing the laptop camera for image capture. It consists of 45k images from 15 users, containing a limited set of gaze labels, specifically only frontal gaze data.

\textbf{EyeDiap:} This dataset is collected in a controlled laboratory setting, where participants are asked to gaze at two types of visual targets: screen targets and a 3D floating ball. It includes 16k images from 16 users, with a restricted set of gaze labels, primarily focused on frontal gaze data.

As observed above, inherent distribution shifts exist between datasets, including variations in image distributions (\textit{e.g.}, collection environments, resolution, and head pose) and gaze label ranges. These disparities make it challenging to merge multiple datasets, highlighting the need for research in cross-dataset gaze estimation. From the perspective of method design, due to distribution shifts among datasets, maintaining separate models for each dataset is often necessary in practice. This motivates our two-stage approach. In the first stage, each single-dataset branch is trained separately. Considering the broad label distribution and non-stationary gazing processes within each dataset, multiple local regressors are employed for each individual dataset, enhancing the performance of with-in domains. In the second stage, building upon the multiple single-dataset branches retained in the production environment, the model enables rapid generalization to new dataset combinations.

\section{Overview of Multi-Source DG}
In this section, we provide a detailed introduction of the two Multi-source Domain Generalization methods, GMDG~\cite{tan2024rethinking} and BNE~\cite{segu2023batch}, which are chosen for performance comparison with our approach.

\textbf{GMDG:} This method is a recent and highly representative approach that introduces a novel and general learning objective, aiming to interpret and analyze the core principles underlying most existing strategies in multi-domain generalization (mDG). Unlike traditional domain generalization methods, which primarily focus on classification tasks, GMDG offers a more generalized framework capable of addressing a wide range of tasks, including regression tasks like ours.

The core of GMDG is a general learning objective that combines four optimization goals, so the loss function is designed as a weighted sum of these four terms, each targeting a specific aspect of the optimization process:

\begin{itemize}
    \item \textbf{GAim1:} This term encourages alignment across domains by minimizing the entropy of the joint distribution of features, \( P(\varphi(X), \psi(Y)) \). The loss for this term is: 
    \begin{equation}
        L_{A1} = \sum_{i=1}^n \left( \log |\Sigma_i| + \|\bar{\mu} - \mu_i\|^2_{\Sigma_i^{-1}} \right),
    \end{equation}
    where \( \Sigma_i \) is the covariance matrix of the feature distribution in domain \(i\), and \( \mu_i \) is the mean feature vector.
    
    \item \textbf{GAim2:} This term further enhances alignment by minimizing the entropy between different arrangements of features, \( P(\psi(Y), \varphi(X)) \) and \( P(Y, \psi(Y)) \). The corresponding loss is:
    \begin{equation}
        L_{A2} = H(P(\psi(Y), \varphi(X))) + H(P(Y, \psi(Y))),
    \end{equation}
    where \( H \) denotes the entropy of the distribution.
    
    \item \textbf{GReg1:} This regularization term minimizes the Kullback-Leibler (KL) divergence between the joint distribution \( P(\varphi(X), \psi(Y)) \) and a prior distribution \( O \) (ImageNet pretrained ResNet-18), which is introduced to prevent overfitting. The loss for this term is:  
    \begin{equation}
        L_{R1} = D_{\text{KL}}(P(\varphi(X), \psi(Y)) \| O),
    \end{equation}
    where \( D_{\text{KL}} \) is the Kullback-Leibler divergence.
    
    \item \textbf{GReg2:} This term suppresses invalid causal relationships by minimizing the Conditional Feature Shift. The corresponding loss is: 
    \begin{equation}
        L_{R2} = \|\Sigma_{XY} \Sigma_{YY}^{-1} \Sigma_{YX}\|^2,
    \end{equation}
    where \( \Sigma_{XY} \), \( \Sigma_{YY} \), and \( \Sigma_{YX} \) are the covariances between different feature distributions.
\end{itemize}

The overall loss function is the weighted combination of these individual loss terms:
\begin{equation}
\small
L(C, \varphi, \psi) = v_{A1} L_{A1} + v_{A2} L_{A2} + v_{R1} L_{R1} + v_{R2} L_{R2},
\end{equation}
where we select the loss weights \( v_{A1} = 0.001 \), \( v_{A2} = 1 \), \( v_{R1} = 0.01 \), and \( v_{R2} = 0.0001 \), based on the depth estimation task, which is a similar regression task to ours, as described in the original paper~\cite{tan2024rethinking}.

\textbf{BNE:} This method is a representative approach for ensemble learning-based multi-source domain generalization, which are compared for two reasons: first, our multi-branch evidential fusion approach can also be seen as an ensemble learning method; second, it shares similarities with the related work on cross-dataset training in autonomous driving, which also addresses inherent distribution shifts using shared BN layers \cite{wang2022cross}.

In this approach, Batch Normalization (BN) parameters \((\mu_d, \sigma_d^2)\) are maintained separately for each domain \(d\), while other model parameters are shared across domains. During inference, the final prediction \(p_t\) for sample \(t\) is computed as a weighted combination of predictions \(p_d^t\) from each domain-specific model. The weight \(w_d^t\) is based on the similarity between the test instance statistics \(r_t^l = (\mu_t^l, \sigma_t^l)\) at layer \(l\) and the accumulated domain statistics \(e_d^l = (\mu_d^l, \sigma_d^l)\), with similarity measured using the 2-Wasserstein distance \(\mathcal{W}\):
\begin{equation}
w_d^t = \frac{1}{\sum_{l \in \mathcal{B}} \mathcal{W}(e_d^l, r_t^l)},
\end{equation}
where \(\mathcal{B}\) denotes the set of layers used for the distance computation.

Finally, the prediction \(p_t\) is calculated as the weighted average of the domain-specific predictions:
\begin{equation}
p_t = \frac{\sum_{d \in \mathcal{D}} w_d^t p_d^t}{\sum_{d \in \mathcal{D}} w_d^t},
\end{equation}
where \(p_d^t\) is the prediction from the domain-specific model for domain \(d\), and \(w_d^t\) is the computed weight based on domain similarity.

\begin{figure*}[h]
 \centering
 \includegraphics[width=14cm]{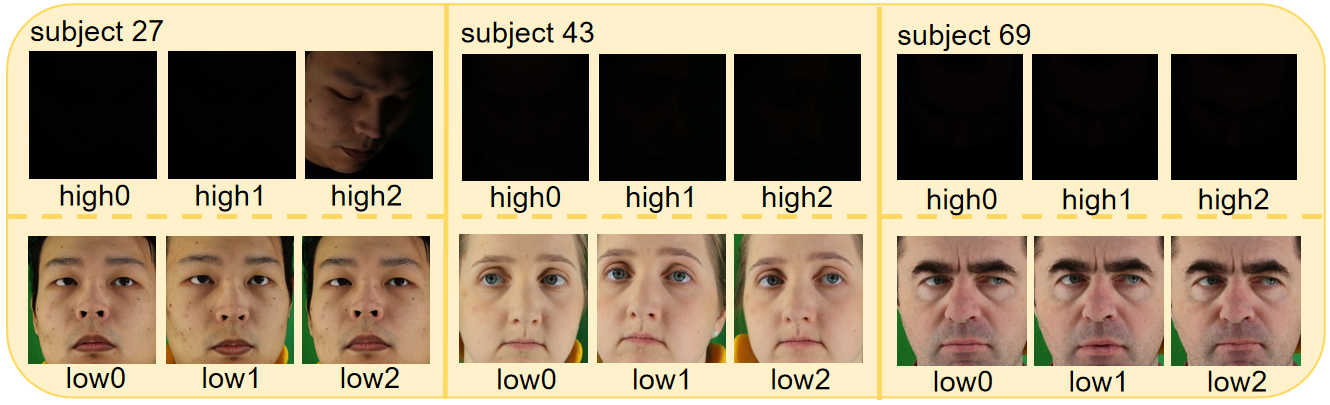}
  \vspace{-8pt} 
  \caption{The aleatoric uncertainty of ETH-Gaze test set.}
  \label{fig:eth_aleatoric_uncertainty}
  \vspace{-5pt} 
\end{figure*}

\begin{figure*}[h]
 \centering
 \includegraphics[width=14cm]{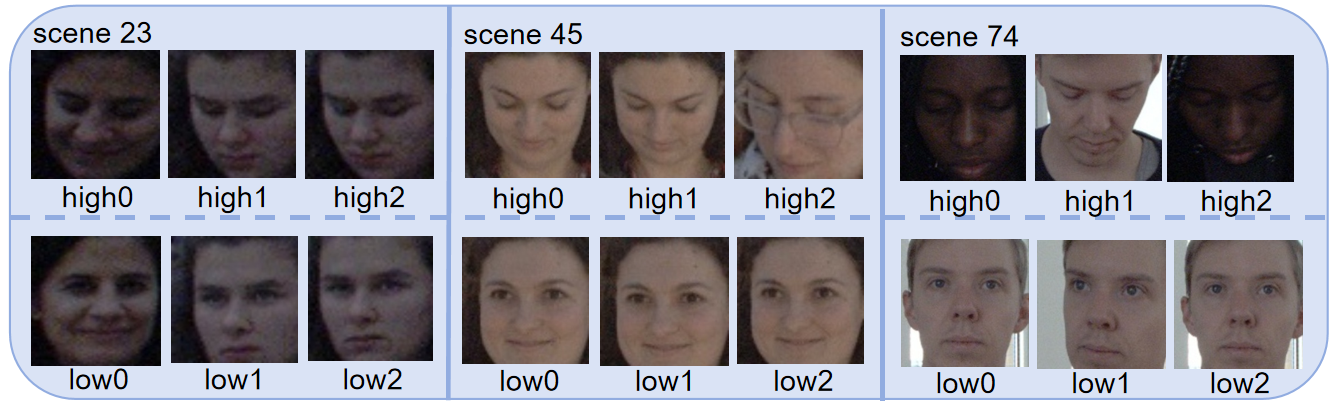}
  \vspace{-8pt} 
  \caption{The aleatoric uncertainty of Gaze360 test set.}
  \label{fig:gaze360_aleatoric_uncertainty}
  \vspace{-5pt} 
\end{figure*}
\section{Aleatoric Uncertainty Visualization} 
 
As shown in Fig.~\ref{fig:eth_aleatoric_uncertainty} and Fig.~\ref{fig:gaze360_aleatoric_uncertainty}, we provide additional visualizations of aleatoric uncertainty for the test sets of two source domain datasets. To facilitate comparison, we visualize the top 3 highest and lowest aleatoric uncertainty values at the user level in the ETH-Gaze dataset and at the scene level in the Gaze360 dataset, since the ETH-Gaze dataset is collected at the individual level, while Gaze360 is collected at the scene level. The results demonstrate that images with occluded, blurred, or invisible eyes exhibit relatively higher uncertainty, showing that our method can accurately assess aleatoric uncertainty.

\end{document}